\documentclass[reprint,aps,nofootinbib,floatfix]{revtex4-2}

\usepackage[T1]{fontenc}
\usepackage{lmodern}
\usepackage{amsmath,amssymb,amsfonts}
\usepackage{graphicx}
\usepackage{booktabs}
\usepackage{microtype}
\usepackage{xcolor}
\usepackage{listings}
\usepackage{tikz}
\usepackage{pgfplots}
\pgfplotsset{compat=1.18}
\usetikzlibrary{shapes,arrows,positioning,fit,backgrounds,shadows,decorations.pathreplacing,calc,arrows.meta}
\usepgfplotslibrary{polar}
\usepackage{hyperref}

\definecolor{NFteal}{HTML}{008080}
\definecolor{NFcyan}{HTML}{00CED1}
\definecolor{NFdark}{HTML}{212529}
\definecolor{NFgray}{HTML}{6c757d}
\definecolor{NFlight}{HTML}{F8F9FA}
\definecolor{NFred}{HTML}{D90429}
\definecolor{NFblue}{HTML}{0077B6}
\definecolor{NForange}{HTML}{FB8500}
\definecolor{NFpurple}{HTML}{6A0DAD}

\tikzset{
    NFbox/.style={
        rectangle, draw=#1, top color=white, bottom color=#1!15,
        rounded corners=3pt, thick, inner sep=6pt, align=center,
        font=\sffamily\scriptsize,
        drop shadow={opacity=0.15, shadow xshift=1.2pt, shadow yshift=-1.2pt}
    },
    NFbox/.default=NFteal,
    NFdatabase/.style={
        cylinder, cylinder uses custom fill, cylinder body fill=#1!10,
        cylinder end fill=#1!30, draw=#1, thick, shape border rotate=90,
        aspect=0.25, align=center, font=\sffamily\scriptsize,
        drop shadow={opacity=0.15, shadow xshift=1.2pt, shadow yshift=-1.2pt}
    },
    NFdatabase/.default=NFteal,
    NFarrow/.style={->, thick, >=Latex, #1},
    NFarrow/.default=NFteal,
    NFlabel/.style={font=\sffamily\tiny\bfseries, text=NFdark, fill=white, inner sep=1.5pt, rounded corners=2pt},
}

\begin{document}

\title{NetForge RL: A Multi-Agent Simulation Environment for Cyber Defense with Durative Actions}

\author{Igor Jankowski}
\affiliation{\texttt{igorjankowwski@gmail.com}}

\begin{abstract}
Training reinforcement-learning agents for cyber defense requires an environment
that reflects the operational setting: noisy, partial observations, several
defenders coordinating across a network, and an adaptive
adversary realized through self-play. We present
\textbf{NetForge~RL}, a multi-agent environment for this setting on
procedurally generated enterprise and operational-technology (OT) networks. A red
agent compromises hosts with partial observability; three zone-split blue agents defend from
synthetic SIEM telemetry, Windows/Sysmon event logs encoded into dense embeddings
rather than a ground-truth state vector. The environment follows the PettingZoo
parallel API with fixed-shape observations and MITRE~ATT\&CK-mapped actions, ships
five scenarios with named difficulty presets and a held-out evaluation split, and
replays deterministically under a seed. A JAX backend vectorizes a reduced
transition core, reaching $2.5\times10^{5}$~environment-steps/s at batch~4096 on
CPU, as a fast surrogate for training-loop iteration. Alongside the environment we provide reference baselines
(scripted, a JAX IPPO trainer, and a self-play tournament), six diagnostic probes
that each measure one defensive skill, and an
evaluation runner reporting 95\% confidence intervals. We describe the
reproducibility engineering behind the environment and include a responsible-use
statement.
\end{abstract}

\maketitle

\section{Introduction}
Cyber defense is a sequential decision problem under partial observation, and a
multi-agent one: several defenders act across separate network zones against an
attacker that adapts~\cite{lowe2017multi}. It also exercises aspects of
RL that toy benchmarks omit: observations are partial and noisy, credit is assigned
over long horizons, objectives are asymmetric, and the action space is large and
grounded in real tactics~\cite{applebaum2016intelligent}.

A cyber-RL environment must offer (i)~a standard RL API, (ii)~observations based on
the telemetry a defender actually sees, (iii)~throughput high enough for MARL sample
budgets, and (iv)~reproducible replay under a seed; existing open-source
environments meet some but not all (Section~\ref{sec:related}).

We introduce \textbf{NetForge~RL} to address these gaps (Figure~\ref{fig:loop}),
covering the four requirements (i)--(iv) and adding evaluation tooling on top:
\begin{itemize}
    \item \textbf{(i) A standard multi-agent API.} The PettingZoo~\cite{terry2021pettingzoo}
    parallel interface with Gymnasium~\cite{towers2024gymnasium} spaces, fixed-shape
    observations, and per-agent action masks, checked by PettingZoo's
    \texttt{parallel\_api\_test}.
    \item \textbf{(ii) Synthetic SOC telemetry.} Blue agents read SIEM logs
    (Windows/Sysmon XML) encoded into an embedding; blue agents never see a
    ground-truth array. A stochastic benign-traffic generator adds background noise so
    alert volume alone is uninformative.
    \item \textbf{(iii) A vectorized backend.} A JAX~\cite{bradbury2018jax} kernel runs a
    reduced transition core under \texttt{jax.vmap}/\texttt{jit}, verified against a
    NumPy reference; it is a throughput surrogate with a coarser observation than the
    Python engine (Section~\ref{sec:jax-backend}), and we report environment-steps and
    agent-steps separately.
    \item \textbf{(iv) Reproducibility.} Per-episode RNGs, reseeded telemetry, and
    deterministic timestamps make rewards and observations replay under a seed,
    guarded by a golden-trajectory test.
    \item \textbf{Evaluation tooling.} Scripted, JAX IPPO, and self-play
    baselines reported with 95\% confidence intervals, and six probes that score
    distinct defensive capabilities.
\end{itemize}

Our contribution is the environment and its evaluation tooling; training reference
RL policies to convergence across all scenarios is future work
(Section~\ref{sec:future}). Section~\ref{sec:limitations} states the environment's
scope and what it does not yet model.

\begin{figure*}[t]
\centering
\begin{tikzpicture}[node distance=0.9cm and 1.15cm]
    \node[NFbox=NFdark] (agents) {Agents\\\tiny red + 3 blue zones};
    \node[NFbox=NFblue, right=of agents] (reg) {Action registry\\\tiny instantiate, mask,\\\tiny cost, duration};
    \node[NFbox=NFpurple, right=of reg] (queue) {Event queue\\\tiny resolve at $t{+}d$};
    \node[NFbox=NFteal, right=of queue] (resolve) {Conflict resolver\\\tiny defensive priority};
    \node[NFdatabase=NFgray, right=of resolve] (state) {Global\\state};
    \node[NFbox=NForange, below=of state] (siem) {SIEM logger\\+ benign traffic\\\tiny Sysmon/Win events};
    \node[NFbox=NFteal, below=of agents] (obs) {Observations\\+ rewards};
    \node[NFbox=NFblue] (enc) at ($(obs)!0.5!(siem)$) {LogEncoder\\\tiny TF-IDF $\to$ SVD};
    \draw[NFarrow=NFdark] (agents) -- node[above,NFlabel]{act} (reg);
    \draw[NFarrow=NFblue] (reg) -- (queue);
    \draw[NFarrow=NFpurple] (queue) -- (resolve);
    \draw[NFarrow=NFteal] (resolve) -- node[above,NFlabel]{apply}(state);
    \draw[NFarrow=NFgray] (state) -- (siem);
    \draw[NFarrow=NForange] (siem) -- node[above,NFlabel]{encode}(enc);
    \draw[NFarrow=NFblue] (enc) -- (obs);
    \draw[NFarrow=NFteal] (obs) -- node[left,NFlabel]{observe}(agents);
\end{tikzpicture}
\caption{The NetForge execution loop. Each \texttt{step} instantiates masked
actions, queues them with per-action durations, resolves simultaneous red/blue
effects, applies state deltas, synthesizes SIEM telemetry (mixed with
benign-traffic noise), encodes it, and returns fixed-shape observations and rewards to
the agents.}
\label{fig:loop}
\end{figure*}

\section{Related Work and Positioning}\label{sec:related}
\textbf{Cyber-defense RL environments.} Open-source cyber-RL environments cluster
into a few families. \emph{Attack-graph abstractions} model an attacker traversing
a network from structured state: NASim~\cite{schwartz2019nasim} is a compact
single-agent POMDP with one-hot features, and CyberBattleSim~\cite{team2021cyberbattlesim}
an abstract single-agent attack graph, since extended to continuous embedding
spaces~\cite{terranova2025ccbs} and to multiple operation
agents~\cite{kunz2023multiagent}. \emph{Operations gyms} such as
CybORG~\cite{standen2021cyborg} and its CAGE challenges~\cite{cage_challenge_2}
provide curated, turn-based red/blue tasks with structured observations.
\emph{Emulation frameworks} such as CSLE~\cite{hammar2023csle} run scenarios on
real virtualized infrastructure for high fidelity, trading away throughput.
Separately, ns3-gym~\cite{gawlowicz2018ns3gym} wraps a network simulator as a Gym
environment for networking research rather than defense.

Few of these combine a native multi-agent API, observations built from log
telemetry rather than ground-truth state, and a batched backend fast enough for
modern sample budgets. NetForge
combines them (Table~\ref{tab:related}). The table reports capabilities, not
quality: each environment has strengths NetForge lacks, for instance CybORG's
curated task designs and CSLE's emulation fidelity.

\begin{table*}[t]
\centering
\caption{Positioning of NetForge~RL relative to representative open-source
cyber-RL environments. ``Multi-agent'' means a native multi-agent RL API;
``SIEM-like obs'' means the observation is encoded log telemetry, not a raw state
vector; ``Vectorized backend'' means a batched/accelerator execution path.}
\label{tab:related}
\begin{tabular}{@{}lccccc@{}}
\toprule
\textbf{Environment} & \textbf{Multi-agent API} & \textbf{SIEM-like obs} & \textbf{Procedural topo.} & \textbf{Vectorized backend} & \textbf{Standard RL API} \\ \midrule
NASim~\cite{schwartz2019nasim}                 & No            & No  & Partial & No  & Gym \\
CyberBattleSim~\cite{team2021cyberbattlesim}   & No            & No  & Yes     & No  & Gym \\
C-CyberBattleSim~\cite{terranova2025ccbs}      & No            & No  & Yes     & No  & Gym \\
CybORG / CAGE~\cite{standen2021cyborg}         & Limited       & No  & Partial & No  & Gym-like \\
CSLE~\cite{hammar2023csle}                     & Limited       & Partial & Yes & No  & Gym-like \\
\textbf{NetForge~RL (ours)}                    & \textbf{Yes}  & \textbf{Yes} & \textbf{Yes} & \textbf{Partial (JAX core)} & \textbf{PettingZoo} \\
\bottomrule
\end{tabular}
\end{table*}

\textbf{Accelerated backends and MARL benchmarks.} Executing the transition
function on accelerators via JAX~\cite{bradbury2018jax} yields large sample
budgets, as Brax shows for rigid-body physics~\cite{freeman2021brax}. NetForge
applies this to a cyber-defense transition core while keeping a Python engine
(event queue, telemetry, command-list effects) for single-environment rollouts.
Our evaluation conventions follow cooperative MARL benchmarks
such as SMAC~\cite{samvelyan2019smac}: fixed-shape observations, per-agent action
masks, and results reported as distributions over seeds.

\section{Environment Design}
NetForge~RL models a red team compromising a generated network and a zone-split
blue team detecting, isolating, and remediating hosts.

\subsection{Agents, Formulation, and API}
The environment is a partially observable, general-sum stochastic game with a
fixed set of four agents: \texttt{red\_operator}, \texttt{blue\_dmz},
\texttt{blue\_internal}, and \texttt{blue\_restricted}.
It implements the PettingZoo \texttt{ParallelEnv} contract: \texttt{reset(seed)}
returns per-agent observations and infos, and \texttt{step(actions)} returns
per-agent observations, rewards, terminations, truncations, and infos. Each blue
agent owns a topological zone (DMZ, internal/corporate, restricted/secure) and
receives SIEM telemetry filtered to that zone plus a shared blue channel; the red
agent operates under partial observability, seeing only hosts it has discovered.
Listing~\ref{lst:api} shows a complete rollout.

\begin{lstlisting}[float=t,caption={A full PettingZoo rollout of NetForge~RL.},label={lst:api}]
import numpy as np
from netforge_rl.environment.parallel_env \
    import NetForgeRLEnv

env = NetForgeRLEnv(
    {"scenario_type": "ransomware",
     "max_ticks": 200})
obs, infos = env.reset(seed=0)
while env.agents:
    actions = {}
    for a in env.agents:
        m = obs[a]["action_mask"]  # 32 types|100 targets
        actions[a] = np.array([
            np.random.choice(np.flatnonzero(m[:32])),
            np.random.choice(np.flatnonzero(m[32:]))])
    obs, rewards, terms, truncs, infos = \
        env.step(actions)
\end{lstlisting}

Each agent's action space is \texttt{MultiDiscrete([32,\,100])}: the first
component selects one of up to 32 action types, the second a target host index
among 100 slots. The \texttt{action\_mask} is the flat concatenation of a 32-bit
type mask and a 100-bit target mask (Table~\ref{tab:obs}). The observation space is a \texttt{Dict} with fixed shapes so
neural policies need no per-topology reshaping (Table~\ref{tab:obs}). We pad the
topology to exactly 100 host slots, of which 15--30 are active, avoiding a
variable-size representation. Padding slots sit on the reserved, non-routable
\texttt{169.254.0.0/16} subnet. The per-step \texttt{action\_mask} zeroes any action
targeting a padding slot, so an agent cannot act on padding and need not learn the
address range. Padding features are constant and shift with the active set under
churn: they cost observation width but leak no episode-specific signal, and we
exclude them from all reported metrics.

\begin{table}[t]
\centering
\caption{Per-agent observation dictionary (fixed shapes). Blue agents receive an
additional \texttt{blue\_comm} channel carrying shared situational awareness.}
\label{tab:obs}
\begin{tabular}{@{}lll@{}}
\toprule
\textbf{Key} & \textbf{Shape} & \textbf{Meaning} \\ \midrule
\texttt{obs}             & $\mathbb{R}^{256}$   & Local host/zone features \\
\texttt{action\_mask}    & $\{0,1\}^{132}$      & Legal types (32) $+$ targets (100) \\
\texttt{siem\_embedding} & $\mathbb{R}^{128}$   & Mean of recent zone SIEM logs \\
\texttt{adj\_matrix}     & $\mathbb{R}^{10000}$ & Flattened $100{\times}100$ reachability \\
\texttt{delta\_t}        & $\mathbb{R}^{1}$     & Normalized time since last step \\
\texttt{blue\_comm}      & $\mathbb{R}^{100}$   & Shared blue awareness (blue only) \\
\bottomrule
\end{tabular}
\end{table}

\subsection{Synthetic SIEM Telemetry}
Rather than exposing ground-truth compromise flags, NetForge renders each
consequential interaction as a realistic Windows Event / Sysmon XML record (e.g.,
\texttt{EventID~4624} logon, \texttt{4625} failed logon, Sysmon~1 process
creation, Sysmon~10 LSASS access). A \texttt{SIEMLogger} buffers these into a
rolling window, and a \texttt{LogEncoder} converts them into a dense
$\mathbb{R}^{128}$ embedding: a character $n$-gram TF-IDF vectorizer projected by
Truncated SVD and $L_2$-normalized. Treating each record as character $n$-grams
keeps the encoder agnostic to log schema and portable to real, semi-structured log
streams, at the cost of not exploiting the categorical field structure a
schema-aware encoder could. Blue agents observe the mean embedding of the eight most
recent logs in their zone, forcing log-based reasoning over telemetry
(Figure~\ref{fig:pipeline}).

\begin{figure}[t]
\centering
\resizebox{\columnwidth}{!}{
\begin{tikzpicture}[node distance=0.9cm]
    \node[NFdatabase=NFgray] (xml) {Event XML\\\tiny\texttt{4624}};
    \node[NFbox=NFblue, right=of xml] (tfidf) {LogEncoder\\\tiny TF-IDF$\to$SVD};
    \node[NFbox=NFteal, right=of tfidf] (win) {Zone window\\\tiny last 8 logs};
    \node[NFbox=NForange, right=of win] (obs) {$\texttt{siem\_emb}$\\\tiny$\in\mathbb{R}^{128}$};
    \draw[NFarrow=NFgray] (xml) -- node[above,NFlabel]{ingest}(tfidf);
    \draw[NFarrow=NFblue] (tfidf) -- node[above,NFlabel]{encode}(win);
    \draw[NFarrow=NFteal] (win) -- node[above,NFlabel]{pool}(obs);
\end{tikzpicture}
}
\caption{The SIEM observation pipeline. Actions emit Windows/Sysmon event logs,
which are encoded and zone-filtered into a fixed-size embedding. A stochastic
benign-traffic generator adds background events to the same buffer.}
\label{fig:pipeline}
\end{figure}

\textbf{Benign-traffic generator (false positives).} To prevent the degenerate
policy of isolating every host, a stochastic benign-traffic generator injects
background events (logins, decoy traffic) into the SIEM buffer each tick. Because
these flow through the same pipeline as adversarial events, the defender must learn
to discriminate true indicators from noise rather than thresholding on volume. These
events are written straight to the SIEM buffer as telemetry; they do not pass through
the event queue (Section~\ref{sec:taxonomy}).

\subsection{Action Taxonomy and Durative Timing}\label{sec:taxonomy}
Actions are grounded in the MITRE~ATT\&CK corpus~\cite{strom2018mitre}. Each agent
belongs to exactly one team (\texttt{red} or \texttt{blue}), and a single action
registry exposes every registered action to that team's live agents. The blue
repertoire spans detection, containment, remediation, and deception (monitoring,
endpoint analysis, EDR deployment, honeytokens) alongside the red attack actions;
a reachability test verifies that every registered action is instantiable by some
agent. In total the environment registers 20 red and 16
blue action types (Table~\ref{tab:actions} lists representatives). Each red action
carries its ATT\&CK technique, so an episode reports which techniques the attacker
actually exercised and the fraction of the mapped taxonomy it covered.

\begin{table}[t]
\centering
\caption{Representative actions with MITRE mappings, energy cost, and duration in
ticks. Red gains and pivots privilege; blue detects, contains, and remediates.
Red rows cite ATT\&CK techniques; blue rows cite ATT\&CK Mitigations (M) or
D3FEND defensive techniques (D3-).}
\label{tab:actions}
\resizebox{\columnwidth}{!}{
\begin{tabular}{@{}lllccl@{}}
\toprule
\textbf{Action} & \textbf{Team} & \textbf{ATT\&CK} & \textbf{Cost} & \textbf{Dur.} & \textbf{Effect} \\ \midrule
\texttt{DiscoverNetworkServices} & Red  & T1046      & 2 & 3 & Port scan; enables exploit \\
\texttt{ExploitRemoteService}    & Red  & T1210      & 5 & 5 & RCE (User) \\
\texttt{ExploitEternalBlue}      & Red  & T1210      & 4 & 6 & SMB RCE \\
\texttt{DumpLSASS}               & Red  & T1003.001  & 1 & 2 & Steal cred. tokens \\
\texttt{PassTheTicket}           & Red  & T1550.003  & 1 & 1 & Token-gated pivot \\ \midrule
\texttt{Monitor}                 & Blue & M1047      & 2 & 2 & Alert on elevated hosts \\
\texttt{Analyze}                 & Blue & M1049      & 1 & 1 & Reveal IoCs (needs EDR) \\
\texttt{IsolateHost}             & Blue & M1030      & 1 & 1 & Sever host, drop sessions \\
\texttt{DeployHoneytoken}        & Blue & D3-DUC     & 5 & 1 & Plant deceptive creds \\
\texttt{RotateKerberos}          & Blue & M1015      &50 & 4 & Global token flush \\
\bottomrule
\end{tabular}
}
\end{table}

Actions have durations and are queued: an action started at tick $t$ with duration
$d$ resolves at $t+d$, and a blue isolation of a host aborts an in-flight red
action targeting it (Figure~\ref{fig:timeline}). An action emits its SIEM telemetry
when it resolves, not when it is submitted, so an in-flight action is not yet
visible; a defender pre-empts an exploit through earlier resolved signals such as
the reconnaissance that precedes it, not by observing the exploit mid-flight. This
gives actions their durative, delayed-effect character while keeping the API a standard
per-step interface (the
normalized time gap is exposed as \texttt{delta\_t}). Each agent may have at
most one action in flight: submitting a new one is blocked until the previous
action resolves, so red and blue bandwidth are both bounded by the same
mechanism. A conflict resolver grants defensive priority when red and blue
effects land on the same tick.

\begin{figure}[t]
\centering
\resizebox{\columnwidth}{!}{%
\begin{tikzpicture}[x=1.0cm, y=1.0cm]
    \draw[->, NFgray, thick] (0,0) -- (7.4,0) node[right,font=\sffamily\small]{tick};
    \foreach \t in {0,1,2,3,4,5,6} \draw[NFgray, thick] (\t,0.08) -- (\t,-0.08) node[below,font=\sffamily\small]{\t};
    \fill[NFred!30, draw=NFred, thick] (0,0.6) rectangle (3,1.5);
    \draw[NFred, dashed, thick] (3,0.6) rectangle (5,1.5);
    \node[font=\sffamily\small, anchor=west, text=NFdark] at (0.05,1.9) {ExploitRemoteService ($d{=}5$)};
    \node[NFred, font=\sffamily\small\bfseries] at (3.5,1.05) {aborted};
    \fill[NFblue!30, draw=NFblue, thick] (2,2.3) rectangle (3,3.2);
    \node[font=\sffamily\small, anchor=west, text=NFdark] at (2.05,3.6) {IsolateHost};
    \draw[NFdark, dashed, thick] (3,-0.3) -- (3,3.9);
    \node[font=\sffamily\small, text=NFdark, anchor=south, align=center] at (5.6,4.0) {isolation cancels\\the in-flight exploit};
    \node[font=\sffamily\small\bfseries, text=NFred, anchor=east] at (-0.15,1.05) {Red};
    \node[font=\sffamily\small\bfseries, text=NFblue, anchor=east] at (-0.15,2.75) {Blue};
\end{tikzpicture}%
}
\caption{Durative timing. A red exploit submitted at $t{=}0$ (duration~5) would
resolve at $t{=}5$, but a blue isolation submitted at $t{=}2$ matures one tick later
and cancels the in-flight exploit at $t{=}3$, so the earlier-maturing action wins and
timing decides the outcome.}
\label{fig:timeline}
\end{figure}

\subsection{Token-Gated Routing}\label{sec:routing}
Reachability is not free: the state engine enforces zone-based routing gated on a
stolen access token. Entering the Secure subnet requires the
red agent to hold an \texttt{Enterprise\_Admin\_Token}, which must be stolen (via
\texttt{DumpLSASS}) and used (via \texttt{PassTheTicket}); a blue
\texttt{RotateKerberos} flushes stolen tokens and revokes that mobility
(Figure~\ref{fig:topo}, abstracted; see Section~\ref{sec:limitations}), so reaching
the Secure zone takes a multi-step attack chain.

\begin{figure}[t]
\centering
\resizebox{\columnwidth}{!}{
\begin{tikzpicture}[node distance=1.0cm]
    \node[circle, draw=NFgray, fill=NFgray!10, thick, minimum size=1.1cm, font=\sffamily\bfseries\tiny] (INET) {Internet};
    \node[NFbox=NFblue, right=of INET] (WEB) {DMZ\\\tiny Web/Mail};
    \node[NFbox=NFteal, right=of WEB] (CORP) {Corporate\\\tiny DC (tokens)};
    \node[NFbox=NFred, right=of CORP] (SEC) {Secure\\\tiny PII/OT [token]};
    \draw[NFarrow=NFgray] (INET) -- node[above,NFlabel]{ingress}(WEB);
    \draw[NFarrow=NFblue] (WEB) -- node[above,NFlabel]{pivot}(CORP);
    \draw[NFarrow=NFred, ultra thick] (CORP) -- node[above,NFlabel]{token}(SEC);
\end{tikzpicture}
}
\caption{Token-gated routing. Traversal into the Secure zone requires a
cryptographic token the red team must steal and can lose to a blue key rotation.}
\label{fig:topo}
\end{figure}

\subsection{Scenarios and Procedural Topologies}
NetForge ships five scenario families that differ in objective and reward
structure: \texttt{ransomware} (mass compromise/encryption),
\texttt{apt\_espionage} (stealthy persistence and exfiltration),
\texttt{cloud\_hybrid} (protect a Secure zone), \texttt{iot\_grid} (protect grid
controllers), and \texttt{ot\_stuxnet} (prevent physical PLC destruction).

Topologies are generated procedurally by \texttt{NetworkGenerator}: 3--5 subnets,
per-host OS/service/CVE/credential profiles, decoys, and an optional OT subnet.
Training draws from arbitrary seeds; setting \texttt{evaluation\_mode=True} draws
from a disjoint held-out seed pool (offset $1000$) that is never seen during
training, giving a clean train/generalization split, and a frozen 20-seed
evaluation suite fixes the reporting set. Difficulty is exposed as three named
presets (\texttt{easy}, \texttt{medium}, \texttt{hard}) that vary network size,
SIEM telemetry delay, and dynamic topology churn (host arrival, DHCP reallocation),
so difficulty and the evaluation split are reproducible and documented. A
curriculum wrapper additionally advances difficulty by a reward threshold for
non-stationarity studies.

\subsection{Reward Function Design}\label{sec:reward}
Every scenario shares one reward shape and differs only in its weights. Each
step, an agent's reward is a shared action-cost penalty (proportional to the
action's energy cost) plus a team-specific sum of weighted terms, each tied to a
discrete state transition, for example a privilege gain, a correct or a
false-positive isolation, a decoy deployment, or a kinetic-destruction event.
Transitions are read through the same \texttt{iter\_host\_deltas} helper as the
metrics (Section~\ref{sec:reproducibility}). The weights themselves are a plain lookup table
per scenario and per team: ransomware blue, for instance, earns $+5$ for
correctly isolating a compromised host and $-2$ for isolating a clean one, while
the OT-safety scenarios (\texttt{ot\_stuxnet}, \texttt{iot\_grid}) score a PLC's
kinetic destruction as $+10^{4}$ to red and $-10^{4}$ to blue. The $\pm10^{4}$
term belongs to the shared shape but its weight is zero in every non-OT scenario,
so catastrophic physical outcomes dominate the return only where a PLC exists.
Because the weights are data, the environment-specification generator
emits them verbatim (Section~\ref{sec:availability}). Raw rewards are not comparable
across scenarios, since each defines its own scale; a $\tanh$-normalized reward in
$[-1,1]$ is provided for cross-scenario reporting.

\section{Reproducibility and Determinism}\label{sec:reproducibility}
Benchmark validity depends on two properties: rewards
must be independent of internal effect encodings, and episodes must replay
identically under a seed.

\textbf{Encoding-invariant rewards.} An action can express its effect as a state
delta or as a command list; rewards and metrics must not depend on which. Both
forms are read through one normalization pass, and a test asserts that the two
encodings of a privilege gain score identically. The JAX backend omits command-list
effects, so the question does not arise there (Section~\ref{sec:jax-backend}).

\textbf{Seeded replay.} Every source of randomness is per-episode and seed-derived:
a per-episode RNG drives all stochastic actions, each stateful component is reseeded
on \texttt{reset(seed)}, and log timestamps advance from a seeded epoch, never the wall
clock. Observations, embeddings, infos, and rewards therefore replay bit-for-bit for
a fixed seed and differ across seeds. A golden-trajectory fingerprint and a
full observable-stream hash guard this, the latter also checking that two
environments interleaved in one process stay independent. The suite (342 tests)
covers the PettingZoo API contract, reward-encoding invariance, seeded replay,
per-scenario termination, and JAX/NumPy parity, and gates continuous integration.

\section{JAX Backend}\label{sec:jax-backend}
The Python engine runs one environment per instance. For the sample budgets MARL
needs, NetForge provides a JAX-vectorized backend that implements a reduced
transition core (host privilege/status, reachability, scenario rewards, action
masks) as pure functions under \texttt{jax.vmap} and \texttt{jax.jit}
(Figure~\ref{fig:jax}), verified against a NumPy reference by a trajectory-level
parity test. The JAX core keeps the Python engine's action-timing semantics: each agent has one
action in flight and cannot submit another until it resolves, and a Blue
\texttt{IsolateHost} that matures on a tick cancels a still-pending Red action on
the same host, with same-tick ties favoring Blue. It does not replicate the Python
engine's event-driven tick-skipping: the vectorized backend advances one tick per
\texttt{step()} rather than jumping to the next event, a fixed-cadence rendition of
the same timing.
In place of the full SIEM text pipeline (synthesized Windows/Sysmon XML through
TF-IDF), a pure, \texttt{jit}-compatible function computes a per-host alert
scalar in $[0,1]$ directly from host state (compromise, privilege, honeytoken,
decoy, EDR coverage). The JAX backend thus does not produce the $\mathbb{R}^{128}$
\texttt{siem\_embedding} of Table~\ref{tab:obs} at all; each host instead carries a
single alert value. The two backends thus agree on host state,
reachability, reward, and action timing, but not on tick cadence or on the shape
and meaning of the observation. The JAX core is therefore for throughput iteration
on the training machinery (debugging a loop, estimating a compute budget), not for
developing observation-dependent policy behavior: policies do not transfer between
backends, and full-fidelity results in this paper use the Python engine.

\begin{figure}[t]
\centering
\resizebox{\columnwidth}{!}{%
\begin{tikzpicture}[node distance=1.3cm]
    \node[NFbox=NFgray, minimum width=1.4cm, minimum height=0.9cm] (s0) {};
    \node[NFbox=NFgray, minimum width=1.4cm, minimum height=0.9cm] (s1) at ([xshift=-6pt,yshift=6pt]s0) {};
    \node[NFbox=NFteal, minimum width=1.4cm, minimum height=0.9cm] (s2) at ([xshift=-6pt,yshift=6pt]s1) {states\\\scriptsize$B{\times}$};
    \node[NFbox=NFblue, right=of s0, minimum height=1.9cm, minimum width=2.6cm] (step) {\texttt{vmap}(step)\\\scriptsize\texttt{jit} on device};
    \node[NFbox=NFgray, right=of step, minimum width=1.4cm, minimum height=0.9cm] (o0) {};
    \node[NFbox=NFgray, minimum width=1.4cm, minimum height=0.9cm] (o1) at ([xshift=-6pt,yshift=6pt]o0) {};
    \node[NFbox=NForange, minimum width=1.4cm, minimum height=0.9cm] (o2) at ([xshift=-6pt,yshift=6pt]o1) {next\\\scriptsize+ reward};
    \draw[NFarrow=NFteal, thick] (s0.east) -- (step.west |- s0.east);
    \draw[NFarrow=NFteal, thick] (s1.east) -- (step.west |- s1.east);
    \draw[NFarrow=NFteal, thick] (s2.east) -- (step.west |- s2.east);
    \draw[NFarrow=NForange, thick] (step.east |- o0.west) -- (o0.west);
    \draw[NFarrow=NForange, thick] (step.east |- o1.west) -- (o1.west);
    \draw[NFarrow=NForange, thick] (step.east |- o2.west) -- (o2.west);
\end{tikzpicture}%
}
\caption{The vectorized backend maps one jit-compiled transition across a batch of
$B$ environment states (stacked cards) with \texttt{jax.vmap}, producing the next
states and rewards without leaving the accelerator.}
\label{fig:jax}
\end{figure}

We report throughput distinguishing \emph{environment-steps} (one batched
transition) from \emph{agent-steps} (environment-steps $\times$ number of agents),
because conflating them inflates numbers by the agent count. On a single
laptop-class CPU (6~cores/12~threads, no GPU; Figure~\ref{fig:throughput}), the JAX core reaches
$2.5\times10^{5}$~environment-steps/s and $1.0\times10^{6}$~agent-steps/s at
batch~4096. The single-instance Python engine, which additionally runs the SIEM
text pipeline, event queue, and command-list effects, sustains
$\approx144$~environment-steps/s on the same machine. Because the two backends do
different work per step, we report each as the cost of its own execution path
rather than a speedup of one over the other; these numbers characterize scaling on
commodity CPU hardware.

\begin{figure}[t]
\centering
\begin{tikzpicture}
\begin{axis}[
    width=\columnwidth, height=4.6cm,
    ybar, ymode=log, log origin=infty,
    bar width=8pt,
    ymin=100, ymax=1e6,
    ylabel={env-steps/s (log)},
    xlabel={batch size $B$},
    ylabel style={font=\sffamily\scriptsize},
    xlabel style={font=\sffamily\scriptsize},
    symbolic x coords={1, 64, 256, 1024, 4096},
    xtick=data,
    xticklabel style={font=\sffamily\scriptsize},
    yticklabel style={font=\sffamily\scriptsize},
    point meta=explicit symbolic,
    nodes near coords, nodes near coords style={font=\sffamily\tiny, rotate=90, anchor=west},
    nodes near coords align={vertical},
    enlarge x limits=0.15,
    tick align=outside,
]
\addplot[draw=NFteal, fill=NFteal!35] coordinates {(1,298)[298] (64,17137)[17k] (256,63283)[63k] (1024,203266)[203k] (4096,249132)[249k]};
\end{axis}
\end{tikzpicture}
\caption{JAX-backend environment-step throughput (log scale) versus batch size on
a single laptop-class CPU (6~cores/12~threads, no GPU), 50 steps with 2 warmup
iterations. Throughput rises from $\approx300$~env-steps/s at $B{=}1$ to
$2.5\times10^{5}$ at $B{=}4096$ as the fixed compilation amortizes over more
environments. The single-instance Python engine, which runs the full SIEM text
pipeline, event queue, and command-list effects, sustains $\approx144$~env-steps/s
on the same machine (Section~\ref{sec:jax-backend}).}
\label{fig:throughput}
\end{figure}

\section{Metrics, Baselines, and Diagnostics}
\textbf{Metrics.} Beyond episodic return, the environment reports operational
metrics in the \texttt{info} dictionary, all computed over active hosts only
(padding excluded): SLA uptime, count of compromised and isolated hosts, mean time
to containment (MTTC), detection rate, total exfiltrated data, a deception-efficacy
ratio (the fraction of red actions that struck a decoy or honeytoken), and ATT\&CK
technique coverage. These let a study distinguish surgical defense from the
degenerate failure mode where an agent minimizes compromise by destroying
network utility.

\textbf{Reference baselines.} We provide mask-aware random and rule-based heuristic
red/blue policies, a scripted kill-chain attacker, a JAX IPPO trainer, and a
benchmark runner (\texttt{python -m benchmarks.run\_benchmark}) that sweeps
scenarios and seeds and reports each metric as a mean with a 95\% confidence
interval, with a train-vs-held-out generalization-gap mode. Because
\texttt{ExploitRemoteService} requires a prior service-discovery action on its
target, a naive heuristic red that fires exploits without reconnaissance never
compromises anything, and the benchmark looks deceptively quiet. A scripted
kill-chain that recons before exploiting and expands its foothold across zones
produces a non-trivial operating point (Table~\ref{tab:baselines}).
We also include a JAX IPPO trainer whose entire rollout (environment step,
observation, policy, and advantage estimation) runs on-device; a
committed 40-iteration run on \texttt{ransomware}
($\approx$246k environment-steps, about a minute on CPU) improves mean blue reward
from $0.13$ to $0.71$. Because this run is on the JAX surrogate with its scalar
alert signal, it confirms only that the mechanics and rewards are learnable there,
not that a policy can master the full $\mathbb{R}^{128}$ SIEM telemetry of the
Python engine. A self-play tournament rates every red and blue policy on a single Elo
ladder via a fractional update: each match's Elo score is Blue's mean SLA
uptime as a continuous share in $[0,1]$ rather than a discrete win/loss, with
Red scored as the complementary share. These baseline implementations serve
as functional demonstrations rather than state-of-the-art benchmark evaluations.

\begin{table}[t]
\centering
\caption{Scripted red baselines against a heuristic blue defender (10 seeds,
150 ticks; active hosts only). A recon-free heuristic attacker never lands an
exploit; the kill-chain attacker compromises hosts and suppresses SLA, giving the
benchmark a non-trivial operating point.}
\label{tab:baselines}
\begin{tabular}{@{}lrr@{}}
\toprule
\textbf{Red policy} & \textbf{Mean compromised} & \textbf{SLA uptime} \\ \midrule
Random                       & $\approx 0.0$ & $\approx 100\%$ \\
Heuristic (no recon)         & $0.0$         & $100\%$ \\
Kill-chain (recon+pivot)     & $2.3$         & $90.4\%$ \\
\bottomrule
\end{tabular}
\end{table}

\textbf{Diagnostics.} Because aggregate return hides \emph{why} a policy succeeds
or fails, NetForge ships a suite of six diagnostic probes, each isolating one
defensive capability: memory (act on a foothold planted at reset), attention
(find one true alert amid heavy noise), temporal reasoning (contain an intrusion
under a delayed SIEM feed), precision (avoid isolating clean hosts), safety
(protect an OT host from kinetic destruction), and generalization (contain a target
while the topology churns). Each probe returns a score in $[0,1]$, and a
\emph{capability card} aggregates the suite across seeds into a per-policy radar
chart (Figure~\ref{fig:radar}), giving a per-capability profile of each policy
instead of one scalar.

Each probe stresses a specific channel and is not a trivial integration check: the
Attention probe plants the one true indicator among equal-severity decoy alerts, the
Temporal probe delays telemetry by a fixed number of ticks, and every probe
penalizes false-positive isolations, not only rewards correct ones. Because the
policies key on the SIEM alert stream, these stresses bite. The reference heuristic
contains clearly-signalled footholds (memory, safety, generalization) but drops to
$0.50$ on attention, where the decoys draw false-positive isolations, and to $0.79$
on temporal, where the delayed feed sets back its response. The random policy scores
near zero except on precision, where inaction is safe, and partly on generalization,
where a chance isolation occasionally lands. The suite thus ranks policies by how
well they act on degraded telemetry.

\begin{figure}[t]
\centering
\begin{tikzpicture}
\begin{polaraxis}[
    width=7.2cm, height=7.2cm,
    xtick={0,60,120,180,240,300},
    xticklabels={Memory,Attention,Temporal,Precision,Safety,General.},
    xticklabel style={font=\sffamily\tiny},
    yticklabel style={font=\sffamily\tiny},
    ymin=0, ymax=1, ytick={0.5,1.0},
    legend style={font=\sffamily\tiny, at={(0.5,-0.16)}, anchor=north, legend columns=2},
]
\addplot+[mark=none, draw=NFteal, thick, fill=NFteal!20, opacity=0.7]
    coordinates {(0,1.0)(60,0.5)(120,0.79)(180,1.0)(240,1.0)(300,1.0)(360,1.0)};
\addlegendentry{Heuristic blue}
\addplot+[mark=none, draw=NFred, thick, fill=NFred!15, opacity=0.6]
    coordinates {(0,0.0)(60,0.0)(120,0.0)(180,0.97)(240,0.0)(300,0.57)(360,0.0)};
\addlegendentry{Random}
\end{polaraxis}
\end{tikzpicture}
\caption{Validating the diagnostic probes (5 seeds): per-capability scores for a
SIEM-reading heuristic and a random defender. The two profiles separate along most
axes (precision excepted, where inaction alone scores well), showing the probes
distinguish a competent policy from a trivial one; the card is meant to profile
trained policies (future work), not to rank these two.}
\label{fig:radar}
\end{figure}

\section{Limitations}\label{sec:limitations}
NetForge is a simulation and its dynamics have \emph{not} been validated against
real network telemetry or live red-team engagements; results do not transfer to
operational systems without further study. Exploit success is abstracted, not
executed. The trainer's demonstration is a single-scenario learning curve, not a
multi-seed, multi-scenario sweep, and the six probes are validated mainly against
scripted policies so far. Several mechanics are deliberate abstractions: the
benign-traffic generator writes to the SIEM buffer directly, so its noise bypasses
the action-duration and preemption mechanics; \texttt{RotateKerberos} revokes
routing to already-held Secure hosts, an abstraction of a real ticket rotation that
invalidates future authentication without severing a live session; the fixed
100-slot encoding pads the $100{\times}100$ reachability matrix to $10^{4}$
mostly-zero entries for 15--30 active hosts, wasteful for dense policies; and the
generator's benign background is a simple template mix that a text encoder can
partly separate by vocabulary, so adversarially-matched noise remains future work
(Section~\ref{sec:future}).

\section{Responsible Use}
NetForge trains attacker as well as defender policies but contains no operational
exploit code and cannot act on real systems: exploit outcomes are CVSS-weighted
probabilities over vulnerability flags, and CVE identifiers are abstract labels, so
trained red policies confer no real-world offensive capability. Users must not
connect the environment to production infrastructure, and a datasheet documents its
generation, intended uses, and limitations.

\section{Future Work}\label{sec:future}
\begin{itemize}
    \item \textbf{Stronger baselines.} A multi-seed, multi-scenario sweep of IPPO,
    MAPPO, and an LLM SOC agent with confidence intervals across all five
    scenarios, replacing the current single-scenario curve, and validating the
    six diagnostic probes on trained and LLM policies rather than scripted
    baseline only.
    \item \textbf{Continuous-time architectures.} \textbf{CT-GMARL}, an ODE-RNN
    graph architecture for multi-agent defense over dynamic network topologies,
    modelling the environment's durative, event-driven timing directly rather than
    through the fixed-cadence, fixed-shape encoding used here.
    \item \textbf{A richer environment.} More scenario and topology variety, an
    expanded diagnostic-probe suite, a frozen versioned benchmark specification,
    graph or sparse-matrix observations to replace the padded adjacency, and
    GPU/TPU-scale throughput alongside the current CPU numbers.
    \item \textbf{Comparison with real networks.} Validating the simulated
    dynamics against real network telemetry or a red-team engagement (closing the
    gap named in Section~\ref{sec:limitations}), and a quantitative related-work
    comparison in place of the qualitative positioning in Table~\ref{tab:related}.
\end{itemize}

\section{Availability}\label{sec:availability}
NetForge~RL is open source at
\url{https://github.com/reforcemind/NetForge_RL}, with documentation at
\url{https://reforcemind.github.io/NetForge_RL/}. The package installs via
\texttt{pip}, exposes the PettingZoo API shown in Listing~\ref{lst:api}, ships the
JAX backend, benchmark scripts, and documentation, and includes the automated test
suite that gates continuous integration. An environment-specification generator
(\texttt{benchmarks/env\_spec.py}) emits a machine-readable description of the
agents, spaces, observability model, termination conditions, and reward
decomposition to support reproducible comparison.

\bibliographystyle{apsrev4-2}
\bibliography{references}

\begin{thebibliography}{16}%
\makeatletter
\providecommand \@ifxundefined [1]{%
 \@ifx{#1\undefined}
}%
\providecommand \@ifnum [1]{%
 \ifnum #1\expandafter \@firstoftwo
 \else \expandafter \@secondoftwo
 \fi
}%
\providecommand \@ifx [1]{%
 \ifx #1\expandafter \@firstoftwo
 \else \expandafter \@secondoftwo
 \fi
}%
\providecommand \natexlab [1]{#1}%
\providecommand \enquote  [1]{``#1''}%
\providecommand \bibnamefont  [1]{#1}%
\providecommand \bibfnamefont [1]{#1}%
\providecommand \citenamefont [1]{#1}%
\providecommand \href@noop [0]{\@secondoftwo}%
\providecommand \href [0]{\begingroup \@sanitize@url \@href}%
\providecommand \@href[1]{\@@startlink{#1}\@@href}%
\providecommand \@@href[1]{\endgroup#1\@@endlink}%
\providecommand \@sanitize@url [0]{\catcode `\\12\catcode `\$12\catcode
  `\&12\catcode `\#12\catcode `\^12\catcode `\_12\catcode `\%12\relax}%
\providecommand \@@startlink[1]{}%
\providecommand \@@endlink[0]{}%
\providecommand \url  [0]{\begingroup\@sanitize@url \@url }%
\providecommand \@url [1]{\endgroup\@href {#1}{\urlprefix }}%
\providecommand \urlprefix  [0]{URL }%
\providecommand \Eprint [0]{\href }%
\providecommand \doibase [0]{https://doi.org/}%
\providecommand \selectlanguage [0]{\@gobble}%
\providecommand \bibinfo  [0]{\@secondoftwo}%
\providecommand \bibfield  [0]{\@secondoftwo}%
\providecommand \translation [1]{[#1]}%
\providecommand \BibitemOpen [0]{}%
\providecommand \bibitemStop [0]{}%
\providecommand \bibitemNoStop [0]{.\EOS\space}%
\providecommand \EOS [0]{\spacefactor3000\relax}%
\providecommand \BibitemShut  [1]{\csname bibitem#1\endcsname}%
\let\auto@bib@innerbib\@empty
\bibitem [{\citenamefont {Lowe}\ \emph {et~al.}(2017)\citenamefont {Lowe},
  \citenamefont {Wu}, \citenamefont {Tamar}, \citenamefont {Harb},
  \citenamefont {Abbeel},\ and\ \citenamefont {Mordatch}}]{lowe2017multi}%
  \BibitemOpen
  \bibfield  {author} {\bibinfo {author} {\bibfnamefont {R.}~\bibnamefont
  {Lowe}}, \bibinfo {author} {\bibfnamefont {Y.}~\bibnamefont {Wu}}, \bibinfo
  {author} {\bibfnamefont {A.}~\bibnamefont {Tamar}}, \bibinfo {author}
  {\bibfnamefont {J.}~\bibnamefont {Harb}}, \bibinfo {author} {\bibfnamefont
  {P.}~\bibnamefont {Abbeel}},\ and\ \bibinfo {author} {\bibfnamefont
  {I.}~\bibnamefont {Mordatch}},\ }in\ \href@noop {} {\emph {\bibinfo
  {booktitle} {Advances in Neural Information Processing Systems (NeurIPS)}}},\
  Vol.~\bibinfo {volume} {30}\ (\bibinfo {year} {2017})\BibitemShut {NoStop}%
\bibitem [{\citenamefont {Applebaum}\ \emph {et~al.}(2016)\citenamefont
  {Applebaum}, \citenamefont {Miller}, \citenamefont {Strom}, \citenamefont
  {Korban},\ and\ \citenamefont {Wolf}}]{applebaum2016intelligent}%
  \BibitemOpen
  \bibfield  {author} {\bibinfo {author} {\bibfnamefont {A.}~\bibnamefont
  {Applebaum}}, \bibinfo {author} {\bibfnamefont {D.}~\bibnamefont {Miller}},
  \bibinfo {author} {\bibfnamefont {B.}~\bibnamefont {Strom}}, \bibinfo
  {author} {\bibfnamefont {C.}~\bibnamefont {Korban}},\ and\ \bibinfo {author}
  {\bibfnamefont {R.}~\bibnamefont {Wolf}},\ }in\ \href@noop {} {\emph
  {\bibinfo {booktitle} {Proceedings of the 32nd Annual Conference on Computer
  Security Applications (ACSAC)}}}\ (\bibinfo {year} {2016})\ pp.\ \bibinfo
  {pages} {363--373}\BibitemShut {NoStop}%
\bibitem [{\citenamefont {Terry}\ \emph {et~al.}(2021)\citenamefont {Terry},
  \citenamefont {Black}, \citenamefont {Grammel}, \citenamefont {Jayakumar},
  \citenamefont {Hari}, \citenamefont {Sullivan}, \citenamefont {Santos},
  \citenamefont {Dieffendahl}, \citenamefont {Horsch}, \citenamefont
  {Perez-Vicente} \emph {et~al.}}]{terry2021pettingzoo}%
  \BibitemOpen
  \bibfield  {author} {\bibinfo {author} {\bibfnamefont {J.}~\bibnamefont
  {Terry}}, \bibinfo {author} {\bibfnamefont {B.}~\bibnamefont {Black}},
  \bibinfo {author} {\bibfnamefont {N.}~\bibnamefont {Grammel}}, \bibinfo
  {author} {\bibfnamefont {M.}~\bibnamefont {Jayakumar}}, \bibinfo {author}
  {\bibfnamefont {A.}~\bibnamefont {Hari}}, \bibinfo {author} {\bibfnamefont
  {R.}~\bibnamefont {Sullivan}}, \bibinfo {author} {\bibfnamefont {L.~S.}\
  \bibnamefont {Santos}}, \bibinfo {author} {\bibfnamefont {C.}~\bibnamefont
  {Dieffendahl}}, \bibinfo {author} {\bibfnamefont {C.}~\bibnamefont {Horsch}},
  \bibinfo {author} {\bibfnamefont {R.}~\bibnamefont {Perez-Vicente}}, \emph
  {et~al.},\ }in\ \href@noop {} {\emph {\bibinfo {booktitle} {Advances in
  Neural Information Processing Systems}}},\ Vol.~\bibinfo {volume} {34}\
  (\bibinfo {year} {2021})\ pp.\ \bibinfo {pages} {15032--15043}\BibitemShut
  {NoStop}%
\bibitem [{\citenamefont {Towers}\ \emph {et~al.}(2024)\citenamefont {Towers},
  \citenamefont {Kwiatkowski}, \citenamefont {Terry}, \citenamefont {Balis}
  \emph {et~al.}}]{towers2024gymnasium}%
  \BibitemOpen
  \bibfield  {author} {\bibinfo {author} {\bibfnamefont {M.}~\bibnamefont
  {Towers}}, \bibinfo {author} {\bibfnamefont {A.}~\bibnamefont {Kwiatkowski}},
  \bibinfo {author} {\bibfnamefont {J.}~\bibnamefont {Terry}}, \bibinfo
  {author} {\bibfnamefont {J.~U.}\ \bibnamefont {Balis}}, \emph {et~al.},\
  }\href@noop {} {\bibfield  {journal} {\bibinfo  {journal} {arXiv preprint
  arXiv:2407.17032}\ } (\bibinfo {year} {2024})}\BibitemShut {NoStop}%
\bibitem [{\citenamefont {Bradbury}\ \emph {et~al.}(2018)\citenamefont
  {Bradbury}, \citenamefont {Frostig}, \citenamefont {Hawkins}, \citenamefont
  {Johnson}, \citenamefont {Leary}, \citenamefont {Maclaurin}, \citenamefont
  {Necula}, \citenamefont {Paszke}, \citenamefont {Vander{P}las}, \citenamefont
  {Wanderman-{M}ilne},\ and\ \citenamefont {Zhang}}]{bradbury2018jax}%
  \BibitemOpen
  \bibfield  {author} {\bibinfo {author} {\bibfnamefont {J.}~\bibnamefont
  {Bradbury}}, \bibinfo {author} {\bibfnamefont {R.}~\bibnamefont {Frostig}},
  \bibinfo {author} {\bibfnamefont {P.}~\bibnamefont {Hawkins}}, \bibinfo
  {author} {\bibfnamefont {M.~J.}\ \bibnamefont {Johnson}}, \bibinfo {author}
  {\bibfnamefont {C.}~\bibnamefont {Leary}}, \bibinfo {author} {\bibfnamefont
  {D.}~\bibnamefont {Maclaurin}}, \bibinfo {author} {\bibfnamefont
  {G.}~\bibnamefont {Necula}}, \bibinfo {author} {\bibfnamefont
  {A.}~\bibnamefont {Paszke}}, \bibinfo {author} {\bibfnamefont
  {J.}~\bibnamefont {Vander{P}las}}, \bibinfo {author} {\bibfnamefont
  {S.}~\bibnamefont {Wanderman-{M}ilne}},\ and\ \bibinfo {author}
  {\bibfnamefont {Q.}~\bibnamefont {Zhang}},\ }\href@noop {} {\bibinfo {title}
  {{JAX}: composable transformations of {P}ython+{N}um{P}y programs}},\
  \bibinfo {howpublished} {\url{http://github.com/jax-ml/jax}} (\bibinfo {year}
  {2018})\BibitemShut {NoStop}%
\bibitem [{\citenamefont {Schwartz}\ and\ \citenamefont
  {Kurniawati}(2019)}]{schwartz2019nasim}%
  \BibitemOpen
  \bibfield  {author} {\bibinfo {author} {\bibfnamefont {J.}~\bibnamefont
  {Schwartz}}\ and\ \bibinfo {author} {\bibfnamefont {H.}~\bibnamefont
  {Kurniawati}},\ }\href@noop {} {\bibinfo {title} {{NASim}: Network attack
  simulator}},\ \bibinfo {howpublished}
  {\url{https://networkattacksimulator.readthedocs.io/}} (\bibinfo {year}
  {2019})\BibitemShut {NoStop}%
\bibitem [{\citenamefont {{Microsoft Defender Research
  Team}}(2021)}]{team2021cyberbattlesim}%
  \BibitemOpen
  \bibfield  {author} {\bibinfo {author} {\bibnamefont {{Microsoft Defender
  Research Team}}},\ }\href@noop {} {\bibinfo {title} {Cyberbattlesim}},\
  \bibinfo {howpublished} {\url{https://github.com/microsoft/CyberBattleSim}}
  (\bibinfo {year} {2021})\BibitemShut {NoStop}%
\bibitem [{\citenamefont {Terranova}\ \emph {et~al.}(2025)\citenamefont
  {Terranova}, \citenamefont {Lahmadi},\ and\ \citenamefont
  {Chrisment}}]{terranova2025ccbs}%
  \BibitemOpen
  \bibfield  {author} {\bibinfo {author} {\bibfnamefont {F.}~\bibnamefont
  {Terranova}}, \bibinfo {author} {\bibfnamefont {A.}~\bibnamefont {Lahmadi}},\
  and\ \bibinfo {author} {\bibfnamefont {I.}~\bibnamefont {Chrisment}},\ }in\
  \href@noop {} {\emph {\bibinfo {booktitle} {International Symposium on
  Research in Attacks, Intrusions and Defenses (RAID)}}}\ (\bibinfo {year}
  {2025})\BibitemShut {NoStop}%
\bibitem [{\citenamefont {Kunz}\ \emph {et~al.}(2023)\citenamefont {Kunz},
  \citenamefont {Fisher}, \citenamefont {La~Novara-Gsell}, \citenamefont
  {Nguyen},\ and\ \citenamefont {Li}}]{kunz2023multiagent}%
  \BibitemOpen
  \bibfield  {author} {\bibinfo {author} {\bibfnamefont {T.}~\bibnamefont
  {Kunz}}, \bibinfo {author} {\bibfnamefont {C.}~\bibnamefont {Fisher}},
  \bibinfo {author} {\bibfnamefont {J.}~\bibnamefont {La~Novara-Gsell}},
  \bibinfo {author} {\bibfnamefont {C.}~\bibnamefont {Nguyen}},\ and\ \bibinfo
  {author} {\bibfnamefont {L.}~\bibnamefont {Li}},\ }in\ \href@noop {} {\emph
  {\bibinfo {booktitle} {International Conference on Computational Science and
  Computational Intelligence (CSCI)}}}\ (\bibinfo {year} {2023})\BibitemShut
  {NoStop}%
\bibitem [{\citenamefont {Standen}\ \emph {et~al.}(2021)\citenamefont
  {Standen}, \citenamefont {Bowman}, \citenamefont {Richer} \emph
  {et~al.}}]{standen2021cyborg}%
  \BibitemOpen
  \bibfield  {author} {\bibinfo {author} {\bibfnamefont {M.}~\bibnamefont
  {Standen}}, \bibinfo {author} {\bibfnamefont {D.}~\bibnamefont {Bowman}},
  \bibinfo {author} {\bibfnamefont {J.}~\bibnamefont {Richer}}, \emph
  {et~al.},\ }\href@noop {} {\bibfield  {journal} {\bibinfo  {journal} {arXiv
  preprint arXiv:2108.09118}\ } (\bibinfo {year} {2021})}\BibitemShut {NoStop}%
\bibitem [{\citenamefont {{TTCP CAGE Working Group}}(2022)}]{cage_challenge_2}%
  \BibitemOpen
  \bibfield  {author} {\bibinfo {author} {\bibnamefont {{TTCP CAGE Working
  Group}}},\ }\href@noop {} {\emph {\bibinfo {title} {{CAGE} Challenge 2}}},\
  \bibinfo {type} {Tech. Rep.}\ (\bibinfo  {institution} {TTCP},\ \bibinfo
  {year} {2022})\BibitemShut {NoStop}%
\bibitem [{\citenamefont {Hammar}\ and\ \citenamefont
  {Stadler}(2023)}]{hammar2023csle}%
  \BibitemOpen
  \bibfield  {author} {\bibinfo {author} {\bibfnamefont {K.}~\bibnamefont
  {Hammar}}\ and\ \bibinfo {author} {\bibfnamefont {R.}~\bibnamefont
  {Stadler}},\ }\href@noop {} {\bibinfo {title} {{CSLE}: A framework for
  building self-learning cyber-security systems}},\ \bibinfo {howpublished}
  {\url{https://github.com/Kim-Hammar/csle}} (\bibinfo {year}
  {2023})\BibitemShut {NoStop}%
\bibitem [{\citenamefont {Gaw{\l}owicz}\ and\ \citenamefont
  {Zubow}(2018)}]{gawlowicz2018ns3gym}%
  \BibitemOpen
  \bibfield  {author} {\bibinfo {author} {\bibfnamefont {P.}~\bibnamefont
  {Gaw{\l}owicz}}\ and\ \bibinfo {author} {\bibfnamefont {A.}~\bibnamefont
  {Zubow}},\ }\href@noop {} {\bibfield  {journal} {\bibinfo  {journal} {arXiv
  preprint arXiv:1810.03943}\ } (\bibinfo {year} {2018})}\BibitemShut {NoStop}%
\bibitem [{\citenamefont {Freeman}\ \emph {et~al.}(2021)\citenamefont
  {Freeman}, \citenamefont {Frey}, \citenamefont {Raichuk}, \citenamefont
  {Girgin}, \citenamefont {Mordatch},\ and\ \citenamefont
  {Bachem}}]{freeman2021brax}%
  \BibitemOpen
  \bibfield  {author} {\bibinfo {author} {\bibfnamefont {C.~D.}\ \bibnamefont
  {Freeman}}, \bibinfo {author} {\bibfnamefont {E.}~\bibnamefont {Frey}},
  \bibinfo {author} {\bibfnamefont {A.}~\bibnamefont {Raichuk}}, \bibinfo
  {author} {\bibfnamefont {S.}~\bibnamefont {Girgin}}, \bibinfo {author}
  {\bibfnamefont {I.}~\bibnamefont {Mordatch}},\ and\ \bibinfo {author}
  {\bibfnamefont {O.}~\bibnamefont {Bachem}},\ }\href@noop {} {\bibfield
  {journal} {\bibinfo  {journal} {arXiv preprint arXiv:2106.13281}\ } (\bibinfo
  {year} {2021})}\BibitemShut {NoStop}%
\bibitem [{\citenamefont {Samvelyan}\ \emph {et~al.}(2019)\citenamefont
  {Samvelyan}, \citenamefont {Rashid}, \citenamefont {de~Witt}, \citenamefont
  {Farquhar}, \citenamefont {Nardelli}, \citenamefont {Rudner}, \citenamefont
  {Hung}, \citenamefont {Torr}, \citenamefont {Foerster},\ and\ \citenamefont
  {Whiteson}}]{samvelyan2019smac}%
  \BibitemOpen
  \bibfield  {author} {\bibinfo {author} {\bibfnamefont {M.}~\bibnamefont
  {Samvelyan}}, \bibinfo {author} {\bibfnamefont {T.}~\bibnamefont {Rashid}},
  \bibinfo {author} {\bibfnamefont {C.~S.}\ \bibnamefont {de~Witt}}, \bibinfo
  {author} {\bibfnamefont {G.}~\bibnamefont {Farquhar}}, \bibinfo {author}
  {\bibfnamefont {N.}~\bibnamefont {Nardelli}}, \bibinfo {author}
  {\bibfnamefont {T.~G.~J.}\ \bibnamefont {Rudner}}, \bibinfo {author}
  {\bibfnamefont {C.-M.}\ \bibnamefont {Hung}}, \bibinfo {author}
  {\bibfnamefont {P.~H.~S.}\ \bibnamefont {Torr}}, \bibinfo {author}
  {\bibfnamefont {J.}~\bibnamefont {Foerster}},\ and\ \bibinfo {author}
  {\bibfnamefont {S.}~\bibnamefont {Whiteson}},\ }\href@noop {} {\bibfield
  {journal} {\bibinfo  {journal} {arXiv preprint arXiv:1902.04043}\ } (\bibinfo
  {year} {2019})}\BibitemShut {NoStop}%
\bibitem [{\citenamefont {Strom}\ \emph {et~al.}(2018)\citenamefont {Strom},
  \citenamefont {Applebaum}, \citenamefont {Miller}, \citenamefont {Nickels},
  \citenamefont {Pennington},\ and\ \citenamefont {Thomas}}]{strom2018mitre}%
  \BibitemOpen
  \bibfield  {author} {\bibinfo {author} {\bibfnamefont {B.~E.}\ \bibnamefont
  {Strom}}, \bibinfo {author} {\bibfnamefont {A.}~\bibnamefont {Applebaum}},
  \bibinfo {author} {\bibfnamefont {D.~P.}\ \bibnamefont {Miller}}, \bibinfo
  {author} {\bibfnamefont {K.~C.}\ \bibnamefont {Nickels}}, \bibinfo {author}
  {\bibfnamefont {A.~G.}\ \bibnamefont {Pennington}},\ and\ \bibinfo {author}
  {\bibfnamefont {C.~B.}\ \bibnamefont {Thomas}},\ }\href@noop {} {\emph
  {\bibinfo {title} {{MITRE ATT\&CK}: Design and Philosophy}}},\ \bibinfo
  {type} {Tech. Rep.}\ (\bibinfo  {institution} {The MITRE Corporation},\
  \bibinfo {year} {2018})\BibitemShut {NoStop}%
\end{thebibliography}%

\end{document}